\documentclass{llncs}
\usepackage{makeidx}  
\usepackage{amsmath, amsfonts}
\usepackage{cite}
\usepackage{graphicx}
\usepackage{kotex}
\usepackage{enumitem}
\usepackage{xcolor}
\usepackage{soul}
\usepackage{footnote}

\usepackage{etoolbox} 
\makeatletter
\patchcmd{\ps@headings}{\rlap{\thepage}}{}{}{}
\patchcmd{\ps@headings}{\llap{\thepage}}{}{}{}

\usepackage{color}
\newcommand{\col}{\color{black}}

\makeatother
\begin{document}

\mainmatter              
\title{Design and Identification of Keypoint Patches in Unstructured Environments}
\titlerunning{Keypoint detection, Image-based tracking, visual servoing}  
%
\author{Taewook Park\inst{1} \and Seunghwan Kim\inst{1} \and Hyondong Oh\inst{1}\thanks{Corresponding author.}}

\authorrunning{Park et al.} 
%
\tocauthor{Taewook Park, Seunghwan Kim, Hyondong Oh}

\institute{Ulsan National Institute of Science and Technology \\ Department of Mechanical Engineering \\Ulsan, Republic of Korea\\
\email{\{wook, kevin6960, h.oh\}@unist.ac.kr}}

\maketitle              

\begin{abstract}
Reliable perception of targets is crucial for the stable operation of autonomous robots. A widely preferred method is keypoint identification in an image, as it allows direct mapping from raw images to 2D coordinates, facilitating integration with other algorithms like localization and path planning. In this study, we closely examine the design and identification of keypoint patches in cluttered environments, where factors such as blur and shadows can hinder detection. We propose four simple yet distinct designs that consider various scale, rotation and camera projection using a limited number of pixels. Additionally, we customize the Superpoint network to ensure robust detection under various types of image degradation. The effectiveness of our approach is demonstrated through real-world video tests, highlighting potential for vision-based autonomous systems.

\end{abstract}

\section{Introduction}
    As robots increasingly take on complex tasks in unstructured environments, the demand for robust perception systems continues to grow. Among various perception techniques, vision-based control has emerged as a particularly popular research area because it can capture abundant environmental information with relatively low cost and compact size \cite{chaumette2016visual, wu2022survey}. These attributes make vision-based systems highly suitable for a wide range of robotic applications, from navigation and manipulation to interaction with humans and other robots.
    
    A core challenge in vision-based control is processing raw visual data into useful information for robotic decision-making. This typically involves extracting meaningful features from 2D images, such as object recognition, tracking, and localization \cite{chen2021review, zhu2020deformable, romero2021tracking}. One of the most popular approaches is detecting and describing interest points within the image. Keypoints are distinctive features in an image that remain invariant to changes in scale, rotation, and illumination, making them highly useful for tasks requiring precise localization and robust matching. By utilizing the direct image coordinates of these points, the robot's control system can accurately interpret the position and movement of objects within its environment.
    
    The detection and description of interest points have been extensively studied for over a decade, with methods generally categorized into traditional computer vision techniques and deep learning-based approaches. Classic methods, such as scale invariant feature transform (SIFT), speeded up robust features (SURF), and oriented fast and rotated BRIEF (ORB), have demonstrated strong resilience to image deformations \cite{karami2017image}. To further enhance the robustness of detection and the quality of keypoint correspondences, deep learning-based methods have been introduced \cite{detone2018superpoint, sun2021loftr}. 
    However, when faced with environmental challenges like lighting variations, shadows, and blur, these algorithms tend to experience significant performance degradation as environmental conditions become more challenging.
    
    {\col We observe that numerous vision-based robots, such as industrial manipulators, repeatedly operate in controlled environments, which deal only with visually similar objects in most cases.} Therefore, we propose novel keypoint patch designs and a specialized training pipeline for the keypoint identification network \cite{detone2018superpoint} to further enhance keypoint identification performance under extreme environmental conditions. To achieve this, we introduce four distinct designs that account for various perspective transformations, including patch rotation, translation, scaling, and projection. {\col To adapt the keypoint identification network for our keypoint patches, we customized Superpoint \cite{detone2018superpoint} to localize and describe the keypoints using a carefully synthesized dataset. Superpoint efficiently localizes keypoints and represents them with a 256-dimensional vector in images of any resolution.} Although the proposed method does not require real-world image acquisition, it demonstrates strong detection performance with images captured by real cameras. The robustness of the results is validated through real videos under various conditions and with artificial effects applied.
    
\section{Method} 

\subsection{Design of Distinct Keypoint Patches}

\begin{figure*}[!ht]
		\centering
		\includegraphics[width=0.9\textwidth]{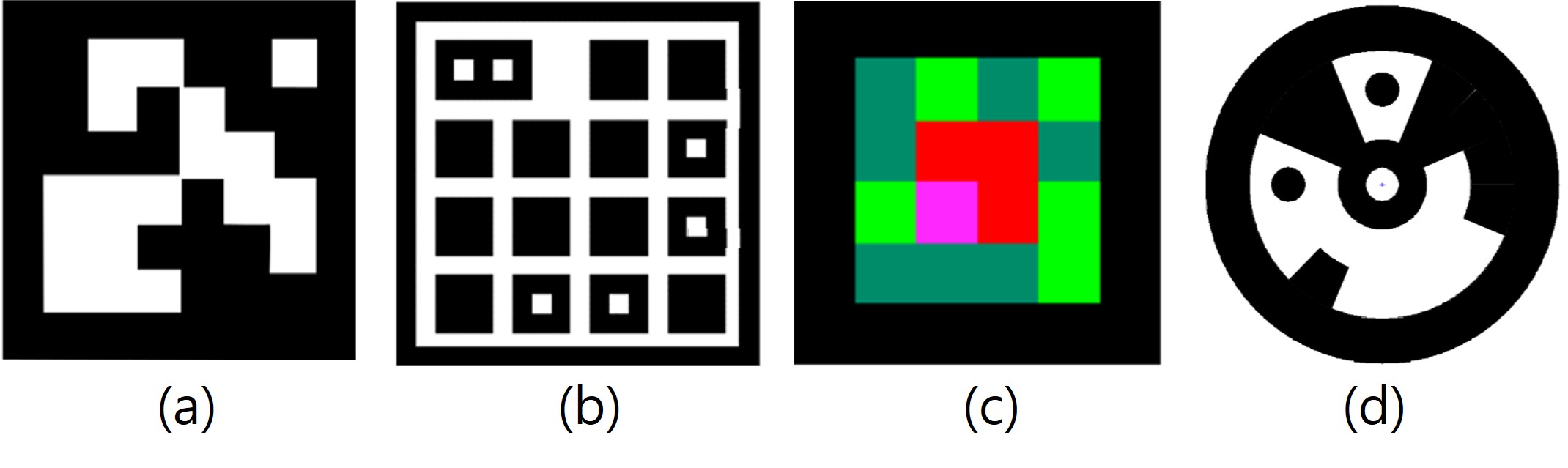}
		\vspace{-5mm}
		\caption{Examples of designs for precise keypoint localization. (a) ArUco3 \cite{romero2018speeded}. (b) Topotag \cite{yu2020topotag}. (c) Chromatag \cite{chen2021autonomous}. (d) Intersense \cite{naimark2002circular}.}
		\vspace{-3mm}
		\label{designs}
	\end{figure*}

	The goal of this section is to design keypoints that are more invariant to perspective transformations, including deformations caused by various scales, rotations, translations, and camera projections. To achieve our objective of defining distinct features, we analyzed existing designs and their limitations in fiducial markers (FMs), which share similar objectives with our research. FMs are widely used for achieving precise and robust localization for robots with  vision sensors. By capturing an image of a predefined marker design, the relative position and orientation between the marker and the camera can be estimated with high accuracy \cite{jurado2023planar}. {\col However, FMs projected in an image can be deformed and distorted depending on the relative pose between the camera and the markers. Therefore, to ensure highly accurate pose estimation of FMs, their designs take into account these geometric transformations.} Figure \ref{designs} shows examples of these designs.
	
	\textbf{\textit{Detection Range and Scale.}} This property is related to the area required to ensure that keypoints are distinguishable from textured background. In FMs, keypoints are typically defined at the boundaries or within the markers. To accurately recognize these keypoints, the markers themselves need to first be identified, which involves detecting unique bit patterns embedded within the marker. As a result, localizing keypoints requires a significant area, limiting the detection range between the camera and the keypoints. Furthermore, defining keypoints by identifying markers presents another challenge across various marker scales. As the scale of the marker increases and its size exceeds the image resolution, the unique bit patterns become unrecognizable. Conversely, at greater distances, these bits may appear as indistinguishable dots, similar to image noise. To address this issue and extend the detection range, keypoints need to be clearly identifiable with minimal area usage. Additionally, defining keypoints using unique bit patterns should be avoided.
	
	\textbf{\textit{Rotation Variance.}} Incorporating rotational uniqueness enhances the versatility of keypoint patches. Circular patterns are preferred over square patterns to achieve this, as square patterns may introduce ambiguity due to the possibility of identical appearances when rotated by 90 degrees. To avoid this duplication and ensure rotational uniqueness, additional information need to be embedded into the pattern, which in turn requires a larger area.
	
	{\col \textbf{\textit{Diverse Lighting Conditions and Background Textures.}} Various environmental factors can complicate keypoint identification. In poorly lit conditions, the RGB intensities near keypoint patches may not be preserved or could even disappear completely. For example, a tree exposed to sunlight may appear either very bright or very dark, depending on the lighting conditions. Moreover, background textures in the environment can sometimes appear as predefined shapes. To address these challenges, keypoint patch designs should be distinct from backgrounds and rely primarily on black and white colors to ensure distinctiveness under varying lighting conditions.}
		
	\begin{figure*}[!ht]
		\centering
		\includegraphics[width=1.0\textwidth]{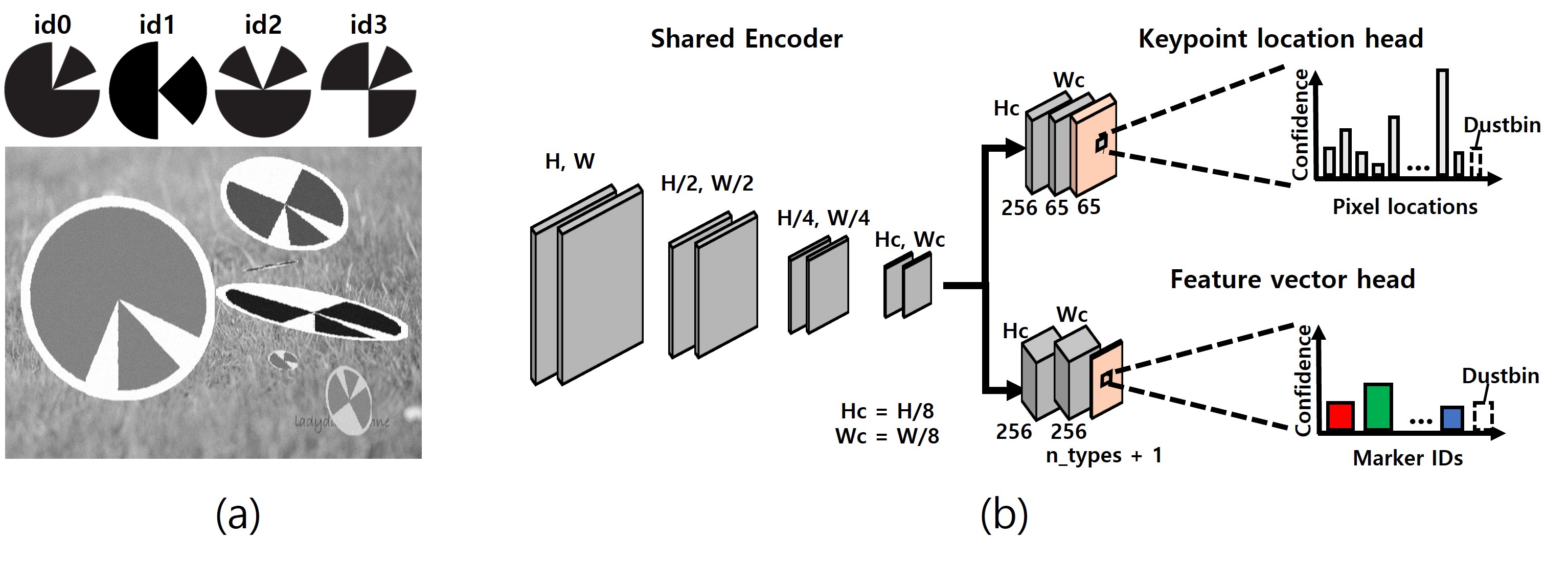}
		\centering
		\vspace{-5mm}
		\caption{Training and inference of Superpoint network customized to proposed keypoint patches. (a) 4 types of keypoint patch are deformed with randomness, and drawn on an image from ImageNet \cite{deng2009imagenet}. (b) The network structure of Superpoint, which predicts 2D keypoint locations and their feature vectors.}
		\vspace{-5mm}
		\label{network}
	\end{figure*}
	
	\textbf{\textit{Proposed design.}} To create distinct keypoint patches suitable for vision-based autonomous robot algorithms, we propose a design concept consisting of semi-circles with shared centers. Four examples are shown in top left at Fig. \ref{network}. The centers of these semi-circles are defined as the keypoints. This design is unique and consistent across various scales, as it can be represented using only a few pixels and is scale-invariant. Its circular shape ensures distinctiveness under orientation changes. Since the keypoints are located within the patch, they are not affected by textured backgrounds. Furthermore, the use of only black and white colors preserves the design's uniqueness under dimming effects and abrupt shadows. By utilizing these keypoint patches, vision-based algorithms can operate more accurately and robustly.

	\subsection{Customized Superpoint Network and Dataset}
	
	The Superpoint network \cite{detone2018superpoint} detects and describes distinct points within an image. It divides the image into an \(8 \times 8\) grid of patches, locating keypoints with associated feature vectors within each patch. However, since the network is designed to work with arbitrary keypoints, the resulting feature vectors are not interpretable and cannot be used to classify known keypoints. Therefore, to enable Superpoint to extract our specific keypoints and classify their types, modifications to the dataset and explicit descriptions of keypoints are necessary. Figure \ref{network} illustrates our training pipeline.

	Our training pipeline utilizes a dataset which contains various real world artifacts. {\col Interest patches are deformed and attached to random backgrounds in a manner similar to that of \cite{zhang2022deeptag}, which added interest features to images. An example is shown in Fig.~\ref{network}(a).} First, random perspective transformations are applied to randomly selected patches. This step mimics random translation, rotation, scale changes and deformations due to the camera projection. To ensure convergence of training, limitation of minimal patch size and degree of deformation is defined. Next, the transformed patches are masked onto an ImageNet dataset \cite{deng2009imagenet}. After that, to consider effects from cluttered environments, various deteriorations such as motion blur, brightness changes, random shadows and even rain effects \cite{buslaev2020albumentations} are applied to an image. 
	\begin{table}[h!]
		\centering
		\caption{PARAMETERS FOR TRAINING SUPERPOINT \cite{detone2018superpoint}}
		\vspace{-2mm}
		\begin{tabular}{|c|c|}
			\hline
			\textbf{Parameter}        & \textbf{Value}    \\ \hline
			Epochs                    & 150               \\ \hline
			$\lambda_{\text{descriptor}}$ & 0.2               \\ \hline
			$\lambda_{\text{d}}$ & \(640 \times 480 / 5 \)                \\ \hline
			$mp$                      & 0.9               \\ \hline
			$mn$                      & 0.2               \\ \hline
			Learning Rate (LR)        & 0.0005            \\ \hline
			LR Decay Rate             & 0.2               \\ \hline
			LR Decay Epochs           & [15, 45]          \\ \hline
			Optimizer                 & Adam              \\ \hline
			Weight Decay              & 0.000001          \\ \hline
		\end{tabular}
		\vspace{0mm}
		\label{tab:training_parameters}
		
	\end{table}
	
	To utilize the pretrained weights and guide the descriptor to predefined vectors, extra layers are attached after Superpoint network as shown in Fig.~\ref{network}(b). One more layer is added to keypoint location head and feature vector head each to seperate pretrained weight and finetuned to our keypoint patches. {\col To guide incomprehensible feature vectors to known quantities, approach of \cite{hu2019deep} is applied. The number of channels in feature vector is set to 5 to account for four keypoint patches as shown in Fig.~\ref{network}(b). The ground truth vector in a patch is one-hot encoded, where an element is set to 1 if the corresponding keypoint patch exists in that \(8 \times 8\) patch. If not, the last element is set to 0. By applying these modifications, a feature vector can be mapped to a predefined vector.}
	
	Detailed parameters for dataset generation and training are listed as follows. At dataset generation phase, 20,000 images are synthesized with \( 640 \times 480 \) resolution. The number of keypoint patches ranges from 0 to 10 in an image. Max ratio difference between long and short axis of keypoint patch ellipse is 0.8. Minimal length of short axis is 10 pixels. The range of intensities of black pixels and white pixels are from 0 to 120 and 180 to 255 respectively. At training phase, only attached layers are trained within 15 epochs. From 16 epoch to 30 epoch, pretrained weights are unfreezed and fine tuned. After 31 epochs, augmentations involving blur, Gaussian noise, random shadows and random rains are applied to the training dataset. Loss functions for the original Superpoint network are used without any modifications. Other parameters used for training are shown in Table \ref{tab:training_parameters}.
	
\section{Experimental Results} \label{sec3}

\subsection{Evaluation Metrics}
To quantitatively assess network prediction performance, we evaluated detection score, feature vector score, and false positives. Given a set of keypoint coordinates \( \mathbf{\hat{X}} = \{\hat{\mathbf{x}}_1, \hat{\mathbf{x}}_2, \dots\, \hat{\mathbf{x}}_n\} \), where each image coordinate \( \hat{\mathbf{x}}_i = \{x_i, y_i\} \) is predicted by the neural network, and \( m \) ground truth keypoints \( \mathbf{X} = \{\mathbf{x}_1, \mathbf{x}_2, \dots, \mathbf{x}_m \} \), detection is considered successful if the following condition is satisfied:

\[
matched(\mathbf{\hat{x}}_i) = \min || \hat{\mathbf{x}}_i - \mathbf{x}_j || \leq \epsilon, 
\]

 \noindent where \(i \in \{1, 2, \dots, n\}\) and \(j \in \{1, 2, \dots, m\} \), and \(\epsilon \) is set to 10\% of the circle radius in keypoint patches, as size of keypoint patches are different in every experiment. \textbf{Detection Score} is defined as the ratio of successful detections to the total number of detections:

\[
\text{Detection Score} = \frac{\sum_{i} 1 \cdot matched(\mathbf{\hat{x}}_i)}{|\mathbf{X}|},
\]

\noindent where \(i \in \{1, 2, \dots, n\}\). Given that \( \mathbf{x} \) is a successful detection, the \textbf{ID Matching Score} checks if the associated feature vector correctly predicts the predefined keypoint patch ID. It is defined by number of correct id predictions over all successful detections. \textbf{Average False Alarm} is the number of detected keypoints that do not correspond to any true keypoints. It is calculated as the average of total number of detections minus the number of successful detections:

\[
\text{Average False Alarm} = \frac{|\hat{\mathbf{X}}| - \textstyle\sum_{i} 1 \cdot matched(\mathbf{\hat{x}}_i)}{n_{image}}.
\]

\noindent Here, \(n_{image} \) denotes the number of images used for evaluation. Since false alarms refer to the number of false positive detections, they have no upper bound, unlike detection scores or ID matching scores, which range from 0 to 1.

\subsection{Results on Validation Dataset}

\begin{figure*}[!ht]
	\centering
	\includegraphics[width=1.0\textwidth]{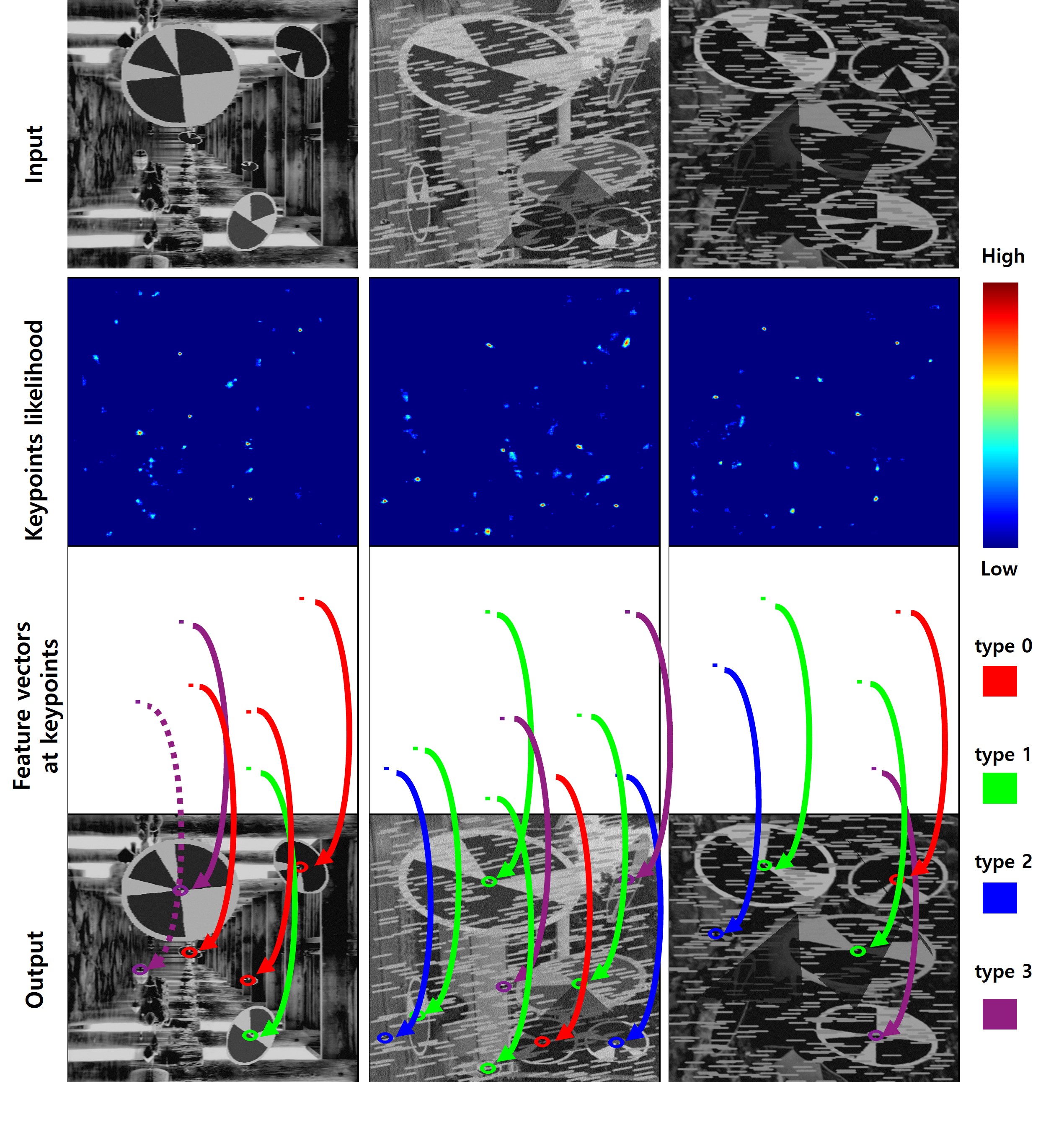}
	\centering
	\vspace{-10mm}
	\caption{Example of network output on validation dataset {\col with image deterioration. 
	\textit{Top}. Input images.
	\textit{Upper}. Confidence map of keypoints at each pixel. \textit{Lower}. Associated patch types to which high-confidence keypoints belong. 
	\textit{Bottom}. The output illustration on the input image highlighting true detections with solid lines, while a dotted line indicating a false positive detection.}}
	\label{syn_validation}
	\vspace{-4mm}
	
\end{figure*}

\indent This section evaluates the performance of the customized Superpoint network on a validation dataset that was not used during training. Total 2,000 images are used. Both qualitative and quantitative results are presented in Table \ref{tab:performance_syn} and Fig. \ref{syn_validation}. The detection and ID matching scores exceed 0.95, demonstrating the network's capability to accurately localize and identify keypoints of interest. {\col Additionally, the false positive detection rates are 0.349 and 0.727 with and without deteriorations, which include blur, Gaussian noise, random shadow and random rain effects. These results} indicate that the network can effectively distinguish between ambiguous background areas and actual keypoint patches.

\subsection{Results on Real World Dataset}
This section assesses the performance of the customized Superpoint network on a real-world dataset. {\col Applying previous detection methods like ORB or original Superpoint may effectively localizes our keypoint patches. However, since they are proposed to detect arbitrary keypoints in an image, they can not identify id of our keypoint patches and thereby can not be fairly compared with proposed customized network. Therefore, we report results using customized Superpoint only.} The network was trained using only synthesized dataset. Images for the evaluation were captured using a Hikrobot MV-CS020-10GC camera, featuring a 4mm focal length lens. The image resolution of camera is \(1624 \times 1240 \).

\begin{table}[!h]
	\caption{VALIDATION RESULTS WITH SYNTHESIZED DATASET}
	\begin{center}
		\begin{tabular}{c|c|c}
			\hline
			\textbf{Metrics}        & \textbf{without deterioration} & \textbf{with deterioration }\\ \hline \hline
			Detection Score  & 0.985                      & 0.950                     \\ \hline
			ID Matching Score & 0.999                      & 0.993                     \\ \hline
			Average False Alarm      & 0.379                      & 0.727                     \\ \hline
		\end{tabular}
	\end{center}
	\label{tab:performance_syn}
	\vspace{-5mm}
\end{table}

To comprehensively evaluate the keypoint patch designs and the customized Superpoint network, six identical patches were printed on a board. The keypoints were arranged in the shape of an equilateral hexagon to facilitate easy validation of detection results, since design assumption enables straightforward identification of false positives and missed detections. Since four different types of keypoint patches were proposed, four markers are printed. To assess keypoint patch deformation, various scales and rotations of the patches were tested. Additionally, the robustness of the Superpoint network was evaluated by introducing artificial image degradations such as blur, dimming, and Gaussian noise.

\subsubsection{Various Scale} \label{sec:various_scale}
\begin{figure*}[!ht]
	\centering
	\includegraphics[width=1.0\textwidth]{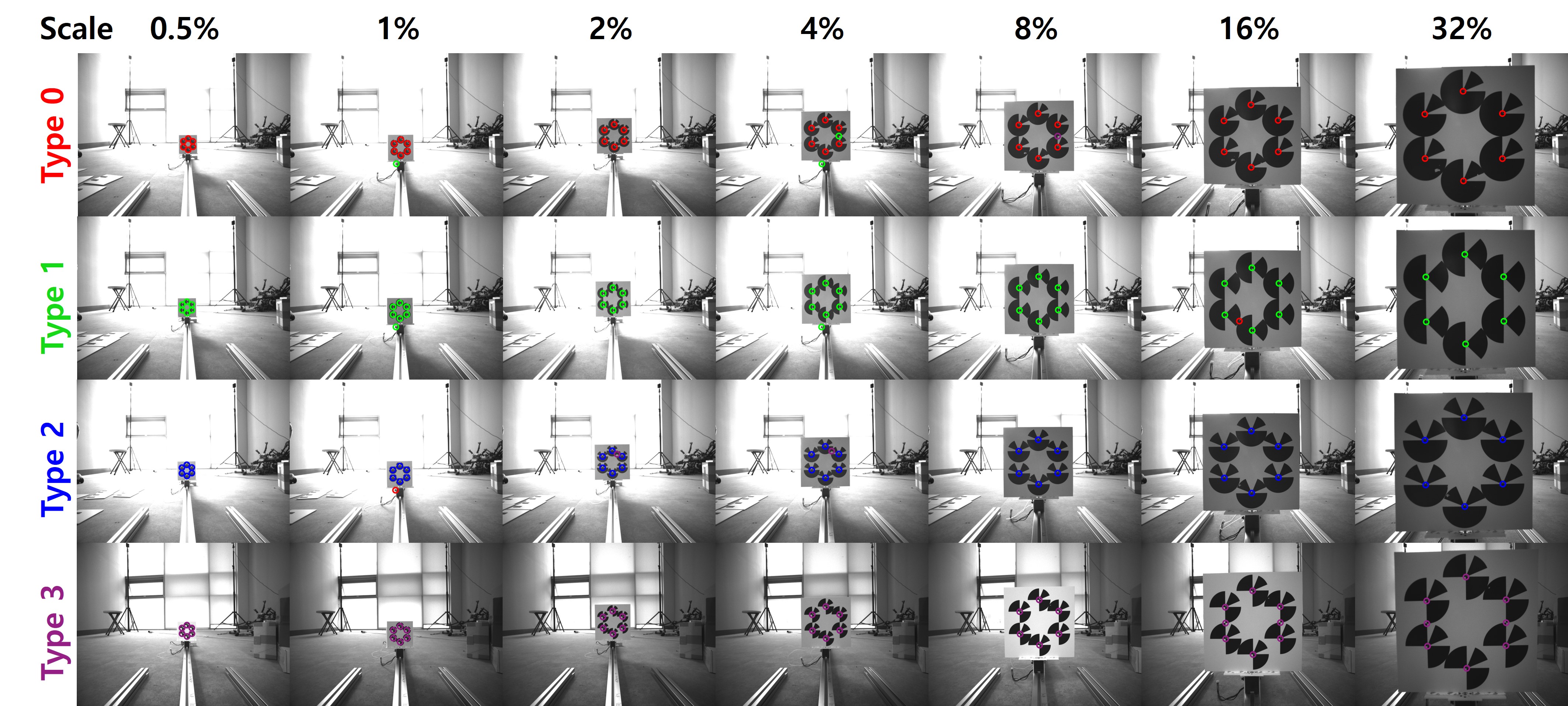}
	\centering
	\vspace{-5mm}
	\caption{Example of network output on real dataset with various scale.}
	\label{qual_scale}
\end{figure*}

This experiment analyzes keypoint patches at various scales. The size of the board occupies between 0.5\% and 32\% of the image. When the marker occupies only 0.5\% of the image, the keypoint patch size becomes 16 pixels. A total of 500 images are used to evaluate each type of keypoint patch at each scale, resulting in a total of 14,000 images for the experiment. The results are presented in Table \ref{tab:performance_scale} and Fig. \ref{qual_scale}. As shown, all keypoint patches are identifiable regardless of their scale. The customized network successfully locates the keypoints and identifies their types. However, due to similar textures in the background and noise in the image, some false positive keypoints are detected.

\subsubsection{Various Rotation}
\begin{figure*}[!ht]
	\centering
	\includegraphics[width=1.0\textwidth]{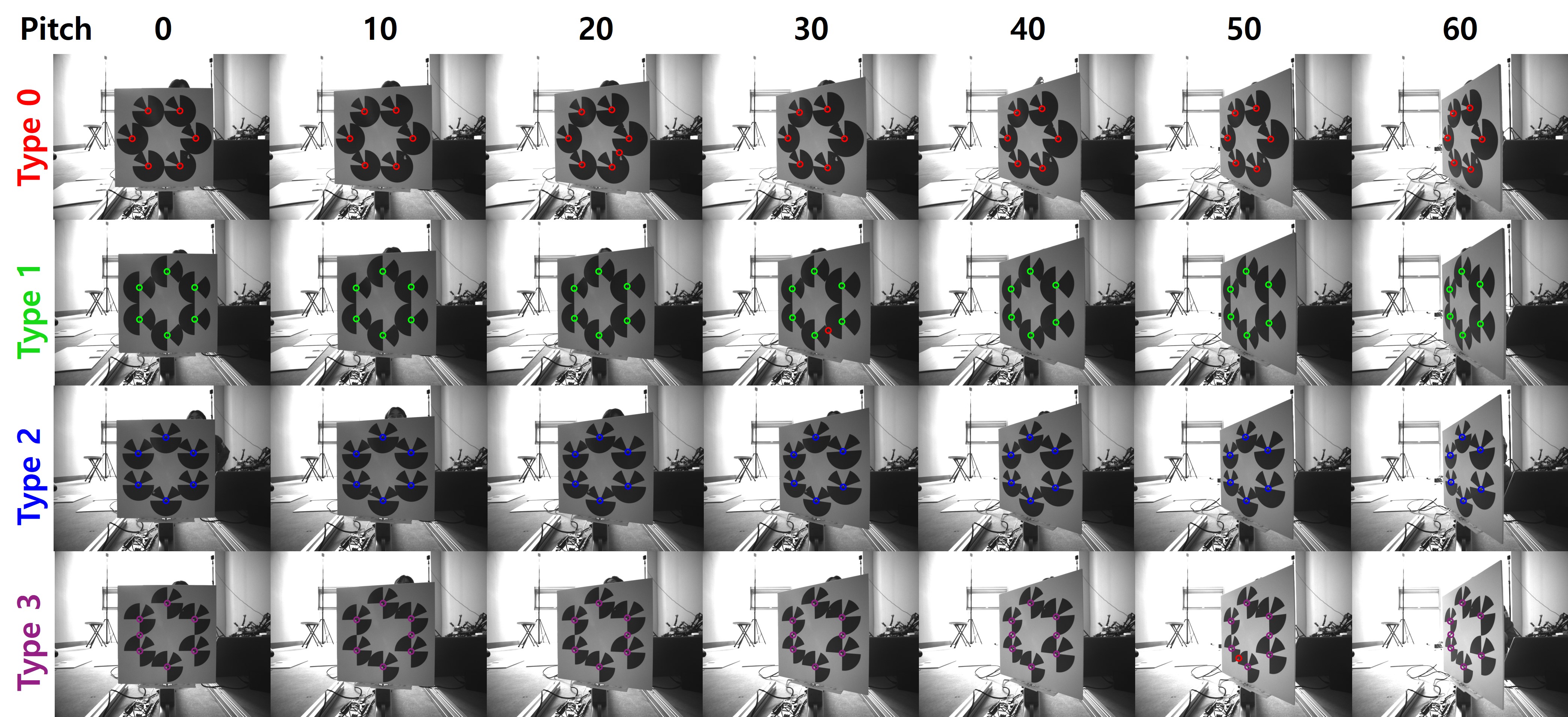}
	\centering
	\vspace{-5mm}
	\caption{Example of network output on real dataset with various pitch angles.}
	\vspace{-2mm} 
	\label{qual_pitch}
\end{figure*}

This experiment examines the performance of detecting rotated keypoint patches. We tilt the pitch angle of the marker relative to the camera coordinate system from 0 degrees to 60 degrees. The occupied area of the board is fixed at 16\% of the image. Similar to Section \ref{sec:various_scale}, 500 images are used for each type and pitch angle, resulting in a total of 14,000 images. The results are presented in Table \ref{tab:performance_pitch} and Fig. \ref{qual_pitch}. All patches are successfully located, and their types are correctly identified regardless of pitch angles.

\subsubsection{Image Deteriorations}
\begin{figure*}[!ht]
	\centering
	\includegraphics[width=0.48\textwidth]{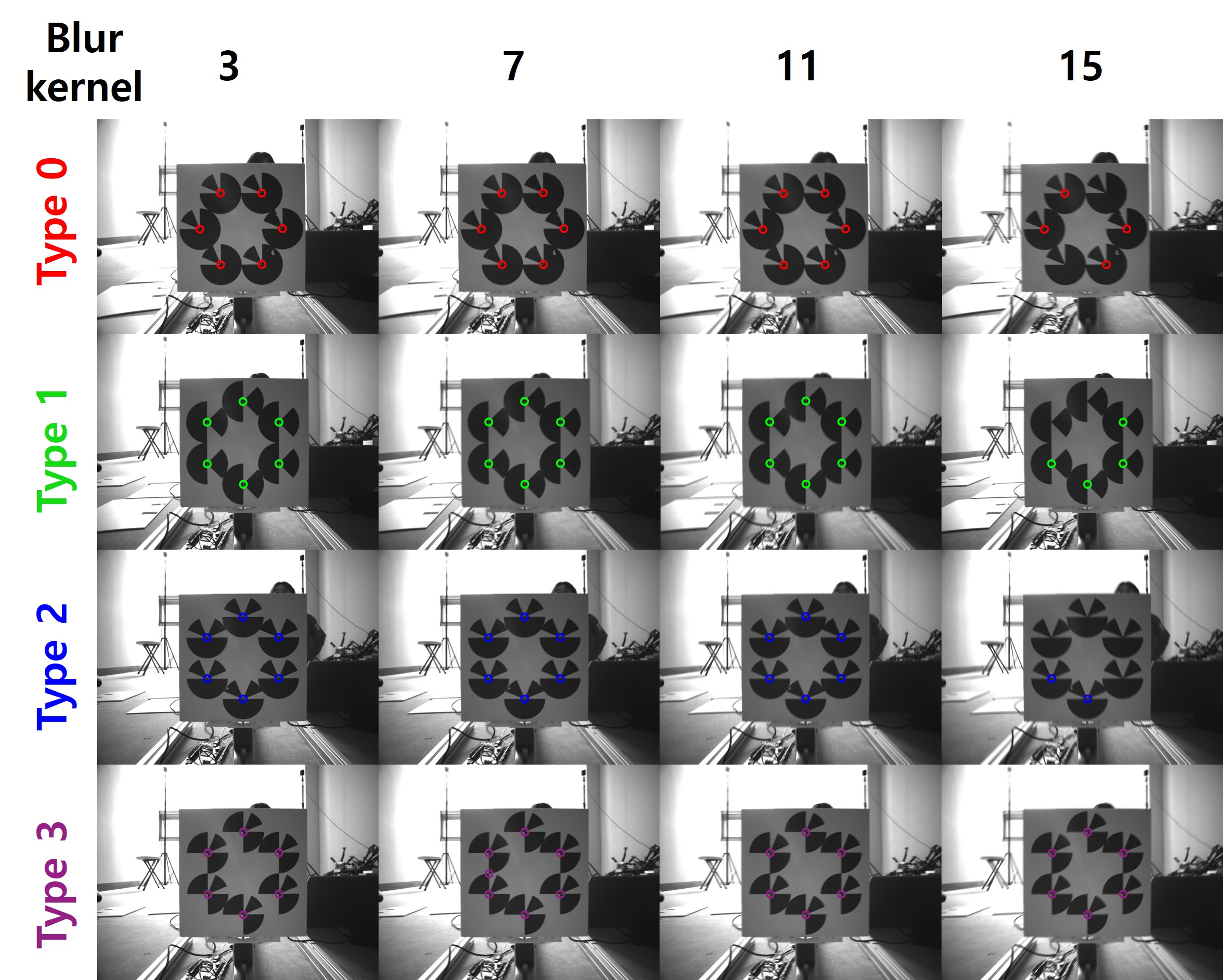}
	\centering
	\vspace{-2mm}
	\caption{Example of network output on real dataset with blur.}
	\vspace{-2mm}
	\label{qual_blur}
\end{figure*}

\begin{figure*}[!ht]
	\centering
	\includegraphics[width=0.48\textwidth]{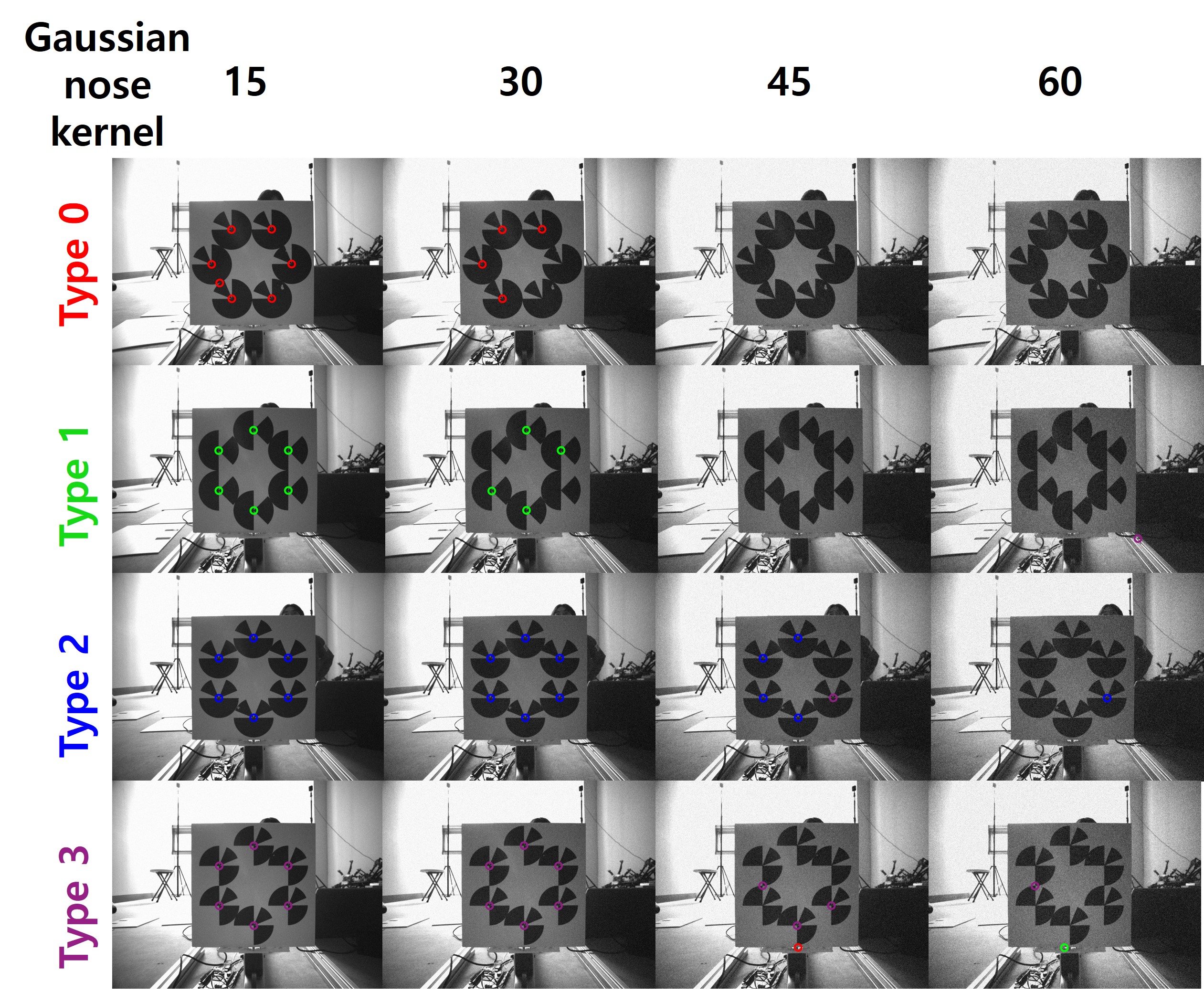}
	\centering
	\vspace{-2mm}
	\caption{Example of network output on real dataset with Gaussian noise.}
	\vspace{-1mm}
	\label{qual_gnoise}
\end{figure*}

This stage examines the proposed design and network customization under artificial conditions such as blur, dimming, and Gaussian noise. The size of the board is fixed at 16\% of the image. Blur is applied with a kernel size ranging from 3 to 15 pixels, causing each pixel to spread out to that size. The dimming effect is applied by multiplying a decreasing factor to all image intensities, calculated by:

\begin{equation}
	\label{eq:dimming_factor}
	f = 0.6^k.
\end{equation}

\noindent where \(k\) ranges from 10 to 40. Gaussian noise is added with a kernel size ranging from 15 to 60 pixels. The results are shown in Tables \ref{tab:performance_blur}\(\sim\)\ref{tab:performance_gaussian_noise} and Figs. \ref{qual_blur}\(\sim\)\ref{qual_gnoise}. Overall, the proposed customized network was able to detect keypoints; however, a high degree of deterioration transformed the proposed patches unidentifiable, leading to detection failures.

\textit{Blur.} The proposed keypoint patches demonstrate robustness to blur. The detection rate exceeds 88\% with a kernel size of 11 pixels but drops to 60\% with a size of 15 pixels. Since proposed design is not preserved under severe blur, the detection rate is lowered. Resizing the image to a smaller size can aid in keypoint detection by reducing the effective blur size. However, resizing also decreases the detection range.

\begin{table}[h!]
	\caption{VALIDATION ON REAL DATASET WITH VARIOUS SCALE}
	\centering
	\begin{tabular}{c|c|c|c|c|c|c|c}
		\hline
		\textbf{Identification Evaluation Metrics} & \textbf{0.5\%} & \textbf{1\%} & \textbf{2\%} & \textbf{4\%} & \textbf{8\%} & \textbf{16\%} & \textbf{32\%} \\ 
		\hline
		\hline
		Detection Score & 1 & 1 & 1 & 1 & 1 & 1 & 0.999 \\ \hline
		ID Matching Score & 1 & 1 & 1 & 1 & 1 & 1 & 1 \\ \hline
		Average False Alarm & 0 & 0.436 & 0.246 & 1.032 & 0.164 & 0.715 & 0.533 \\ \hline
	\end{tabular}
	\label{tab:performance_scale}
	\vspace{-3mm}
\end{table}

\begin{table}[h!]
	\caption{VALIDATION ON REAL DATASET WITH VARIOUS ROTATION}
	\centering
	\begin{tabular}{c|c|c|c|c|c|c|c}
		\hline
		\textbf{Identification Evaluation Metrics} & \textbf{0} & \textbf{10} & \textbf{20} & \textbf{30} & \textbf{40} & \textbf{50} & \textbf{60} \\ \hline \hline
		Detection Score    & 1     & 0.999  & 0.996  & 1     & 1     & 1     & 1     \\ \hline
		ID Matching Score  & 1     & 1      & 1      & 1     & 1     & 1     & 1     \\ \hline
		Average False Alarm & 0.344 & 0.486  & 0.3385 & 0.569 & 0.592 & 0.245 & 0.284 \\ \hline
	\end{tabular}
	\vspace{2mm}

	\label{tab:performance_pitch}
\end{table}

\begin{table}[h!]
	\caption{VALIDATION ON REAL DATASET WITH BLUR}
	\centering
	\begin{tabular}{c|c|c|c|c}
		\hline
		\textbf{Identification Evaluation Metrics}       & \textbf{3} & \textbf{7} & \textbf{11} & \textbf{15} \\ \hline \hline
		Detection Score    & 1        & 0.986  & 0.882  & 0.605  \\ \hline
		ID Matching Score  & 1        & 1      & 1      & 1      \\ \hline
		Average False Alarm & 0.257    & 0.266  & 0.111  & 0.034  \\ \hline
	\end{tabular}
	\label{tab:performance_blur}
	\vspace{20mm}
\end{table}

\begin{table}[h!]
	\vspace{-21mm}
	\caption{VALIDATION ON REAL DATASET WITH DIMMING}
	\centering
	\begin{tabular}{c|c|c|c|c}
		\hline
		\textbf{Identification Evaluation Metrics}       & \textbf{10} & \textbf{20} & \textbf{30} & \textbf{40} \\ \hline \hline
		Detection Score    & 1        & 1        & 0.988    & 0.783  \\ \hline
		ID Matching Score  & 1        & 1        & 1        & 1      \\ \hline
		Average False Alarm & 0.037    & 0        & 0        & 0      \\ \hline
	\end{tabular}
	\label{tab:performance_dimming}
	\vspace{-2mm}
\end{table}

\begin{table}[h!]
	\caption{VALIDATION ON REAL DATASET WITH GAUSSIAN NOISE}
	\centering
	\begin{tabular}{c|c|c|c|c}
		\hline
		\textbf{Identification Evaluation Metrics}       & \textbf{15} & \textbf{30} & \textbf{45} & \textbf{60} \\ \hline \hline
		Detection Score    & 1        & 0.872  & 0.332  & 0.071  \\ \hline
		ID Matching Score  & 1        & 0.998  & 0.892  & 0      \\ \hline
		Average False Alarm & 0.916    & 0.302  & 0.296  & 0.556  \\ \hline
	\end{tabular}
	\label{tab:performance_gaussian_noise}
	\vspace{-2mm}
\end{table}

\begin{figure*}[!ht]
	\centering
	\includegraphics[width=0.48\textwidth]{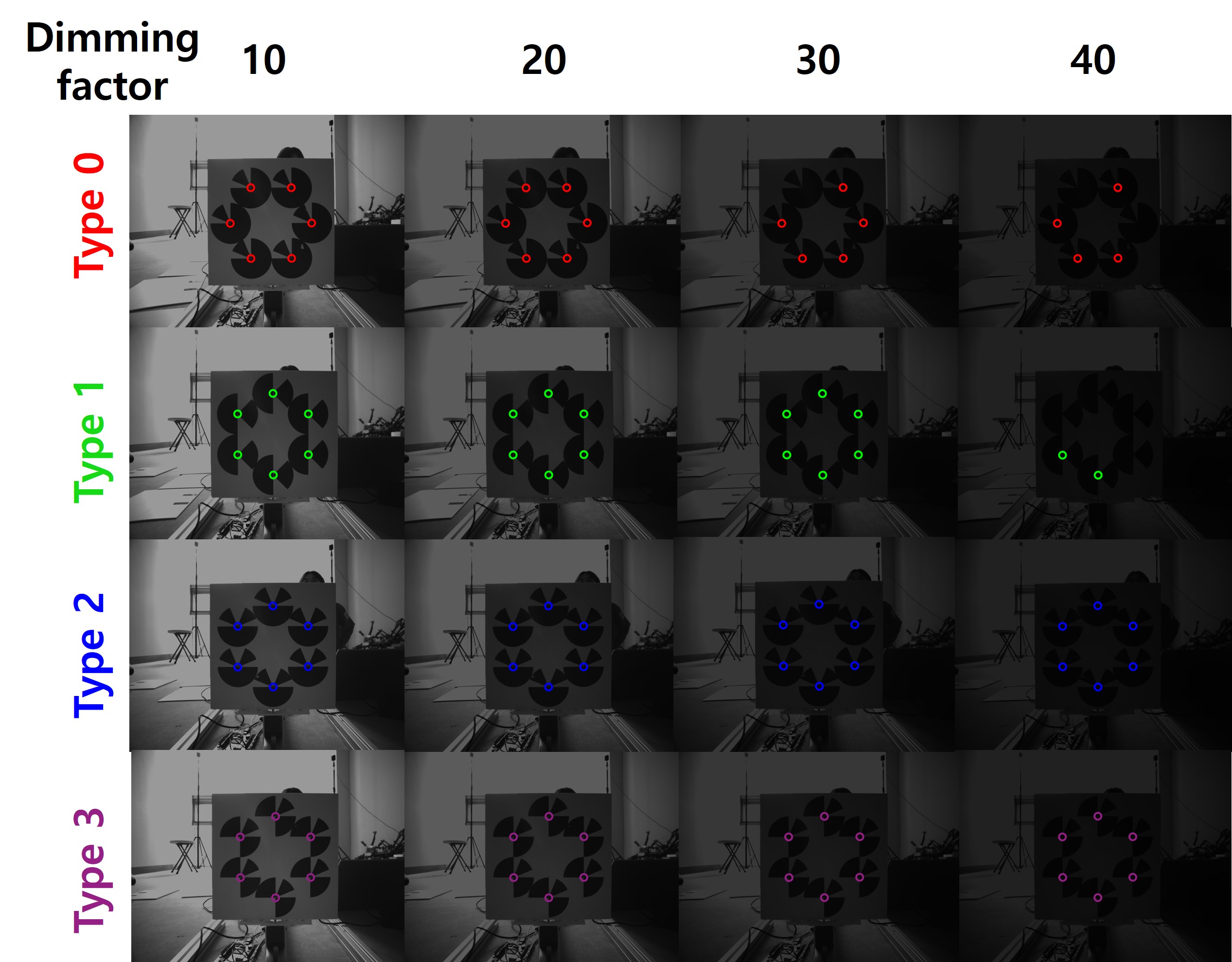}
	\centering
	\vspace{-2mm}
	\caption{Example of network output on real dataset with dimming.}
	\vspace{-1mm}
	\label{qual_dimming}
\end{figure*}

\textit{Dimming.} The proposed method exhibits resilience to poorly lit conditions, with more than 78\% of keypoints detected in the worst-case scenario.

\textit{Gaussian Noise.} The proposed method yields unsatisfactory results under severe levels of Gaussian noise. Distinguishing small keypoint patches from high levels of Gaussian noise is challenging, leading to the Superpoint network's failure to accurately distinguish them. However, such extreme noise levels are typically mitigated by modern camera technologies in general real-world applications. In summary, the proposed method is sufficiently applicable to real-world problems.

\section{Conclusions and Future work}
Reliable detection of interest points is essential for vision-based autonomous systems. To this end, we designed four types of keypoints along with a deep neural network-based detector. The proposed designs are simple yet informative, making them identifiable despite various shape deformations. The customized Superpoint network \cite{detone2018superpoint}, trained on a fully synthesized dataset, is capable of handling abrupt image deteriorations such as blur and dimming.
The application of these proposed patches can enhance the stability of vision-based robotic tasks, particularly in constrained and specialized environments like indoor navigation or manipulator control.

Notably, since the proposed keypoint patches require only a tiny area for identification, they can be detected under significant occlusion. Such situations frequently occur in the real-world due to factors like shadows, lens flare from sunlight, image fading, or the presence of dynamic objects. Therefore, leveraging this characteristic can provide a solution for keypoint detection in highly cluttered and dynamic environments, a longstanding challenge for vision-based autonomous systems.

\section*{Acknowledgments}
This research was supported by National Research Foundation of Korea (NRF) grant funded by the Korea government (MSIT) (2023R1A2C2003130) and Basic Science Research Program through the National Research Foundation of Korea (NRF) funded by the Ministry of Education (2020R1A6A1A03040570).

\bibliographystyle{splncs}
\bibliography{references.bib}
\end{document}